\documentclass[sigconf]{acmart}
\usepackage{multirow}
\usepackage{adjustbox}
\usepackage{subfigure}
\usepackage{enumitem}
\usepackage{subcaption}
\usepackage{booktabs}
\usepackage{setspace}
\AtBeginDocument{%
  }

\copyrightyear{2025}
\acmYear{2025}
\setcopyright{acmlicensed}\acmConference[WWW Companion '25]{Companion
Proceedings of the ACM Web Conference 2025}{April 28-May 2, 2025}{Sydney,
NSW, Australia}
\acmBooktitle{Companion Proceedings of the ACM Web Conference 2025 (WWW
Companion '25), April 28-May 2, 2025, Sydney, NSW, Australia}
\acmDOI{10.1145/3701716.3715545}
\acmISBN{979-8-4007-1331-6/2025/04}

\newenvironment{sequation}
{\begin{equation}}{\end{equation}}

\def\model{\texttt{TRACE}}

\begin{document}
\title{\model: Intra-visit Clinical Event Nowcasting via Effective Patient Trajectory Encoding}

\author{Yuyang Liang}
\affiliation{%
  \institution{The Chinese University of Hong Kong, Shenzhen}
  \city{Shenzhen}
  \state{Guangdong}
  \country{China}
}
\email{yuyangliang@link.cuhk.edu.cn}

\author{Yankai Chen}
\affiliation{%
  \institution{Cornell University}
  \city{Ithaca}
  \state{New York}
  \country{Unite States}
}
\email{yankaichen@acm.org}

\author{Yixiang Fang}
\affiliation{%
  \institution{The Chinese University of Hong Kong, Shenzhen}
  \city{Shenzhen}
  \state{Guangdong}
  \country{China}
}
\email{fangyixiang@cuhk.edu.cn}

\author{Laks V. S. Lakshmanan}
\affiliation{%
  \institution{University of British Columbia}
  \city{Vancouver}
  \state{British Columbia}
  \country{Canada}
}
\email{laks@cs.ubc.ca}

\author{Chenhao Ma}
\authornote{Chenhao Ma is the corresponding author.}
\affiliation{%
  \institution{The Chinese University of Hong Kong, Shenzhen}
  \city{Shenzhen}
  \state{Guangdong}
  \country{China}
}
\email{machenhao@cuhk.edu.cn}

\def\cyk{\color{red} }

\renewcommand{\shortauthors}{Yuyang Liang, Yankai Chen, Yixiang Fang, Laks V. S. Lakshmanan \& Chenhao Ma}

\begin{abstract}
  Electronic Health Records (EHR) have become a valuable resource for a wide range of predictive tasks in healthcare. However, existing approaches have largely focused on \textit{inter-visit} event predictions, overlooking the importance of \textit{intra-visit} nowcasting, which provides prompt clinical insights during an ongoing patient visit.
  To address this gap, we introduce the task of laboratory measurement prediction within a hospital visit. 
  We study the laboratory data that, however, remained underexplored in previous work.
  We propose \model, a Transformer-based model designed for clinical event nowcasting by encoding patient trajectories. \model~ effectively handles long sequences and captures temporal dependencies through a novel timestamp embedding that integrates decay properties and periodic patterns of data. Additionally, we introduce a smoothed mask for denoising, improving the robustness of the model. Experiments on two large-scale electronic health record datasets demonstrate that the proposed model significantly outperforms previous methods, highlighting its potential for improving patient care through more accurate laboratory measurement nowcasting. The code is available at https://github.com/Amehi/TRACE.

\end{abstract}



\maketitle

\begin{figure}[t]
\centering
    \subfigure[]{%
        \includegraphics[ width=0.23\textwidth]{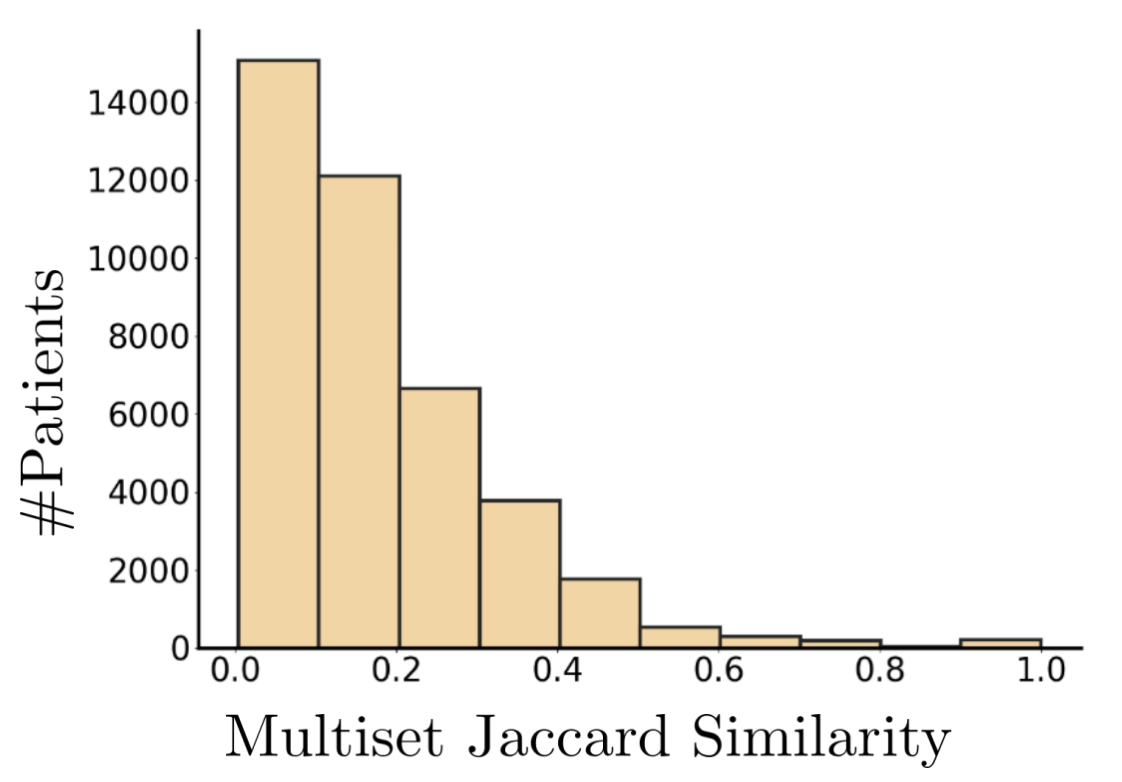}
        \label{fig:distri}}
    \subfigure[]{
        \includegraphics[ width=0.23\textwidth]{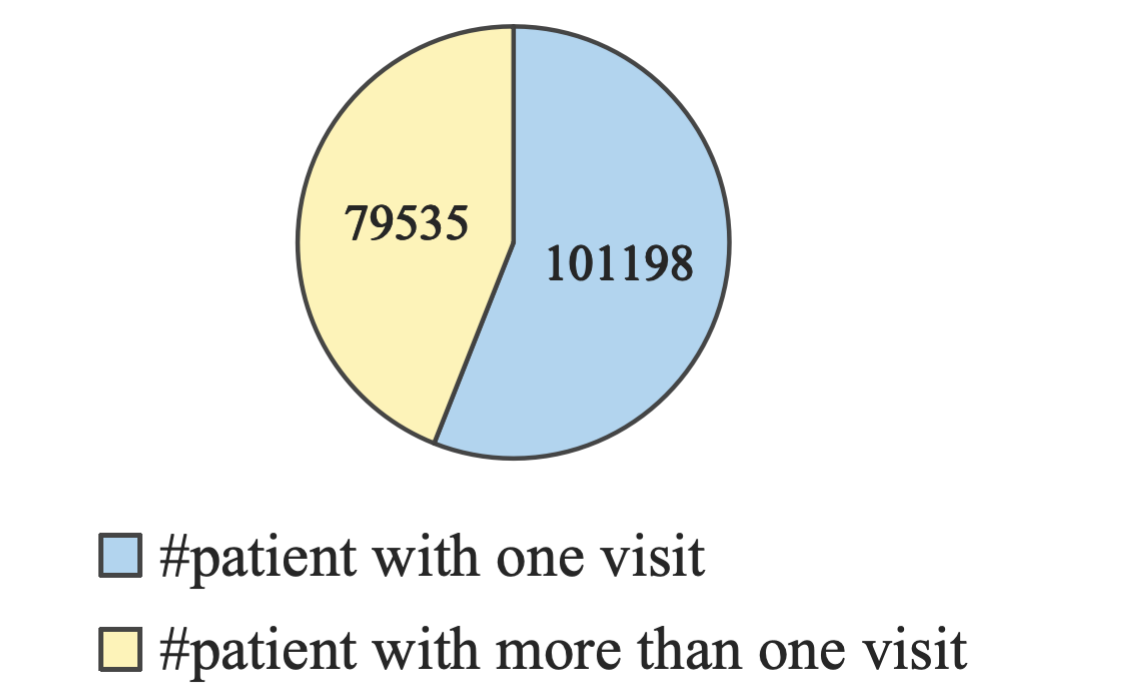}
        \label{fig:prop}{}}
    \vspace{-10pt}
    \caption{Investigation on MIMIC-IV dataset. (a) Jaccard similarity of diagnosis codes between different visits of patients. (b) The proportion of patients with only one visit and patients with multiple visits in the MIMIC-IV dataset.}\label{fig:dataset}
\end{figure}

\section{Introduction}

Electronic Health Records (EHR) provide comprehensive digital records of patients’ medical histories, including diagnoses, medications, procedures, and laboratory results. With the growing complexity of this data, advanced predictive models have emerged to enhance patient outcomes through accurate medical event nowcasting, early risk detection, and personalized treatment, optimizing healthcare delivery.

Traditional EHR predictive methods mainly follow conventional sequential modeling~\cite{wu2023survey,cstar,chen2023star}, to focus on \textit{inter-visit} predictions, nowcasting events after the current visit, such as readmissions \cite{liu_toward_2023} or future diagnoses \cite{ma_kame_2018, choi_gram_2017, gao_camp_2019}. 
While valuable, these models overlook \textit{intra-visit} predictions, which aim to provide real-time insights into a patient’s condition during a single visit. Such tasks may offer more valuable insights into current patient health statuses than merely nowcasting changes between separate visits, as medical information for different diagnoses in multiple visits may not be strongly correlated. 
To illustrate this issue, we investigate the MIMIC-IV dataset, with our observations summarized in Figure \ref{fig:dataset}. 
Figure \ref{fig:distri} shows the Jaccard similarity of diagnosis codes for patients with multiple visits, revealing low similarity scores and relatively weak correlations between visits. Additionally, visit-based models exclude single-visit patient data, as shown in Figure \ref{fig:prop}, leading to significant information loss for over half of the patients in the MIMIC-IV dataset. This loss may impair model performance for newly admitted patients, limiting its real-world effectiveness.


To address these issues and provide timely healthcare, we consider the \textit{intra-visit clinic event nowcasting}. Unlike prior works on inter-visit EHR data, we explore \textit{laboratory measurements in EHR}, such as blood tests and metabolic panels, which provide vital insights into a patient’s current clinical condition. Despite their importance, laboratory data is underutilized in existing models. In this work, we propose the laboratory measurement prediction within a hospital visit, focusing on nowcasting a patient’s clinical state in the next time window based on observed medical events. The detailed task descriptions are provided in Section \ref{task}.
Specifically, We propose a patient \textbf{TR}ajectory encoding \textbf{A}pproach for \textbf{C}linal \textbf{E}vent nowcasting (\model), a novel Transformer-based model tailored for \textit{intra-visit} EHR data analysis. 
To provide time-sensitive patient trajectory encoding, \model~incorporates a customized timestamp embedding that captures decay and periodicity in medical events.
This design allows the model to effectively account for the influence of past events while incorporating the impact of temporal cycles on laboratory test results. \model~leverages a multi-layer Transformer architecture~\cite{10.5555/3295222.3295349}, which has been widely demonstrated to be effective across various applications~\cite{lin2024effective,chen2024deep,chen2022effective,chen2020literature}.
\model~enables it to capture complex temporal dependencies and hierarchical relationships among medical events for more accurate laboratory measurement predictions.
Additionally, we introduce a smoothed mask denoising module that enhances robustness by filtering out less influential information during attention calculation, enabling fine-grained denoising while maintaining differentiability for efficient backpropagation and optimization.
Extensive experiments conducted showing \model’s superior performance, achieving the highest PR-AUC and Precision@k scores on both MIMIC-III and MIMIC-IV datasets.

\section{Task Descriptions}\label{task}

{\textbf{Patient Visit Trajectory.}}
A visit trajectory in EHR for a patient can be represented as a time-ordered sequence of medical events. Formally, a trajectory for patient $p$ during visit $v$ is represented as:
\[
\mathcal{T}_{p,v} = \{ (e_1, t_1), (e_2, t_2), \dots, (e_n, t_n) \},
\]
where $e_i$ denotes the medical code of $i$-th medical event, and $t_i$ is the corresponding timestamp. Each event $e_i$ can encompass various types of medical information such as diagnoses, procedures, medications, or laboratory results. The sequence is ordered by time, i.e., $t_1 \leq t_2 \leq \dots \leq t_n$, reflecting the chronological order in which the events occurred during the patient's visit.

{\textbf{Medical Event.}}
For medication and procedure records, $e_i$ refers to a specific medical concept, such as a drug (e.g., aspirin) or a procedure (e.g., MRI). Laboratory measurements are more complex, with $e_i = (a_i, f_i)$, where $a_i$ is the medical code of test subject and $f_i$ is a flag indicating the result. In MIMIC-III, $f_i$ is binary ($f_i = 0$ for normal, $f_i = 1$ for abnormal), while in MIMIC-IV, $f_i$ provides more details, indicating whether the result is higher, lower, or normal relative to reference bounds. Moreover, Consecutive test records often share the same timestamp $t_i$, reflecting scenarios where multiple tests are conducted simultaneously, such as after a single blood draw. We use the last few records with the same timestamp as the laboratory events to predict.

{\textbf{Laboratory Measurement Prediction.}}
Given a sequence of past events in a visit trajectory
$\mathcal{T}_{p,v} = \{ (e_1, t_1), (e_2, t_2), \dots, (e_k, t_k) \}$ 
up to time $t_k$, the objective is to forecast a set of upcoming laboratory test events $\mathcal{T}_{p,v}'= \{ (e_{k+1}, t_{k+1}), \dots, (e_{k+m}, t_{k+m}) \}$ for $t_{k+1} > t_k$ and $t_{k+1} = \dots = t_{k+m}$.

\section{\model}\label{method}

    

\begin{figure}[tb] 
\centering
\includegraphics[ width=0.45\textwidth]{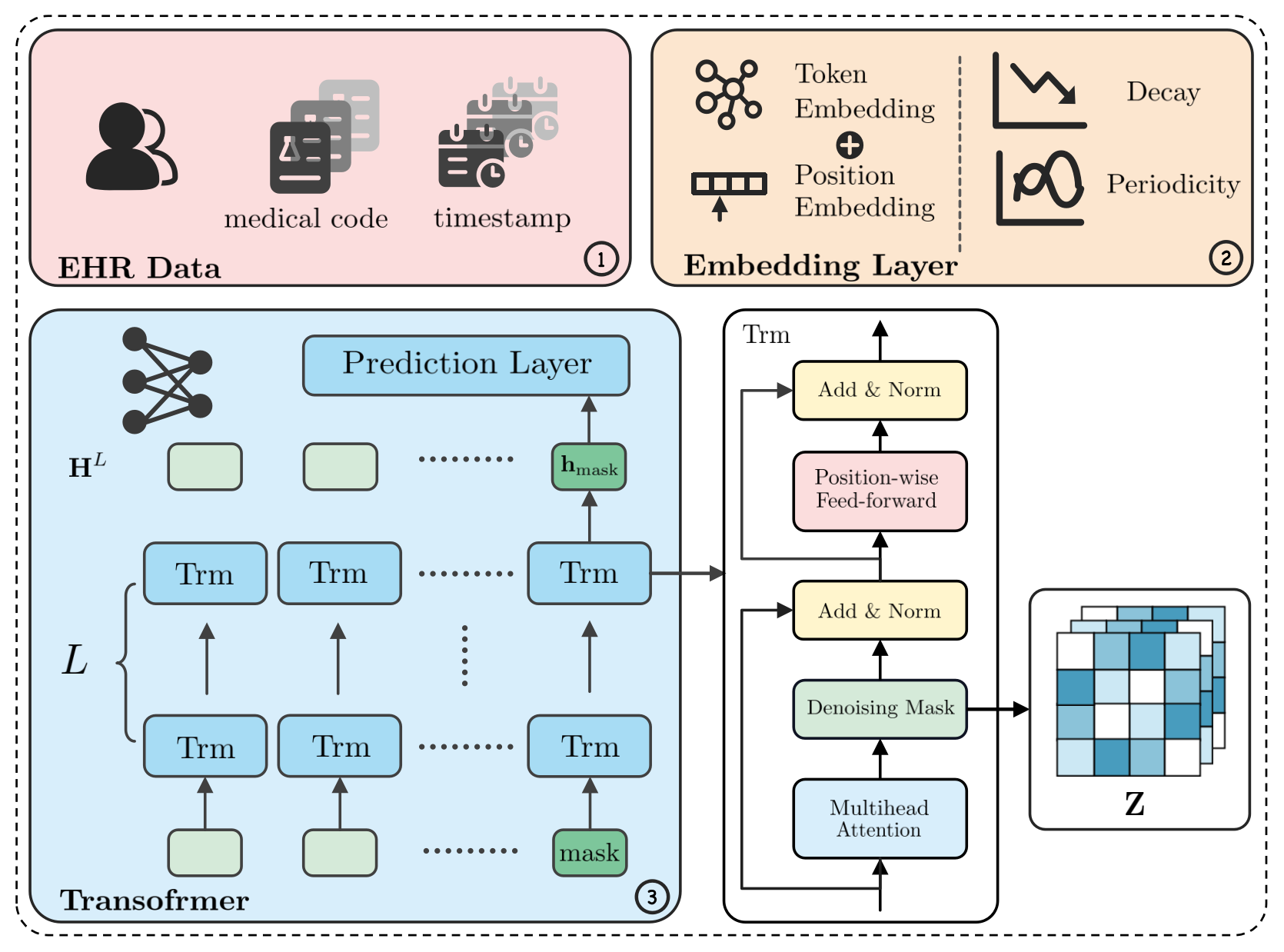}
\caption{The \model ~model workflow.}
\label{fig:model}
\end{figure}

\begin{table*}[thb]
\setlength{\tabcolsep}{0.62em}
    \centering
    \caption{Experiment result for laboratory measurement prediction task. The threshold is set to 0.5 for F1-score.} %
    \begin{adjustbox}{max width=\textwidth}
        \begin{tabular}{@{}l  l| c c c  c  c  c  c  |c@{} }
        \toprule
        \textbf{Datasets} & \textbf{Metric} & \textbf{Transformer} & \textbf{RETAIN} & \textbf{Dipole} & \textbf{HiTANET} & \textbf{ConCare} & \textbf{GRASP} & \textbf{PPN} & \textbf{\model}\\
        \midrule
        \multirow{4}{*}{MIMIC-III} 
        & F1-score & 0.5705 \(\pm\) 0.0196 & \underline{0.7439} \(\pm\) 0.0047& 0.5795 \(\pm\) 0.0020& 0.5669 \(\pm\) 0.0049& 0.5884 \(\pm\) 0.0247& 0.5734 \(\pm\) 0.0096 & 0.6441 \(\pm\) 0.0167& \textbf{0.7858} \(\pm\) 0.0055 \\
        & PR-AUC & 0.3468 \(\pm\) 0.0006 & 0.4993 \(\pm\) 0.0013 & 0.2745 \(\pm\) 0.0081 & \underline{0.5350} \(\pm\) 0.0025 & 0.3931 \(\pm\) 0.0117 & 0.2546 \(\pm\) 0.0414 & 0.4478 \(\pm\) 0.0044  & \textbf{0.6476} \(\pm\) 0.0042\\
        & Precision@5 & 0.4663 \(\pm\) 0.0009 & 0.5469 \(\pm\) 0.0018 & 0.4374 \(\pm\) 0.0136 & \underline{0.5800} \(\pm\) 0.0034 & 0.4888 \(\pm\) 0.0091 & 0.3263 \(\pm\) 0.0250 &  0.5215 \(\pm\) 0.0011& \textbf{0.6075} \(\pm\) 0.0045\\
        & NDCG@5 & 0.4948 \(\pm\) 0.0068 & 0.6086 \(\pm\) 0.0021 & 0.4664 \(\pm\) 0.0142 & \underline{0.6494} \(\pm\) 0.0045 & 0.5336 \(\pm\) 0.0164 & 0.3426 \(\pm\) 0.0346 & 0.5818 \(\pm\) 0.0017 & \textbf{0.6872} \(\pm\) 0.0067\\
         \midrule
        \multirow{4}{*}{MIMIC-IV}
        & F1-score & 0.6570 \(\pm\) 0.0354 & 0.7526 \(\pm\) 0.1452 & 0.6276 \(\pm\) 0.0020 & 0.5473 \(\pm\) 0.0010 & 0.7539 \(\pm\) 0.0061 & 0.6541 \(\pm\) 0.0552 & \underline{0.8000} \(\pm\) 0.0071 & \textbf{0.8535} \(\pm\) 0.0062\\
        & PR-AUC & 0.4969 \(\pm\) 0.0004 & \underline{0.6893} \(\pm\) 0.0027 & 0.3962 \(\pm\) 0.0227 & 0.6720 \(\pm\) 0.0028 &  0.5103 \(\pm\) 0.0095 & 0.4400 \(\pm\) 0.0772 & 0.6394 \(\pm\) 0.0119 & \textbf{0.7777} \(\pm\) 0.0021\\
        & Precision@5 & 0.6391 \(\pm\) 0.0001 & 0.7252 \(\pm\) 0.0021 & 0.5139 \(\pm\) 0.0753 & \underline{0.7531} \(\pm\) 0.0032 & 0.6395 \(\pm\) 0.0018 & 0.4872 \(\pm\) 0.0303 & 0.7099 \(\pm\) 0.0062 & \textbf{0.7566} \(\pm\) 0.0023\\
        & NDCG@5 & 0.6413 \(\pm\) 0.0001 & 0.7462 \(\pm\) 0.0023 & 0.5185 \(\pm\) 0.0766 & \underline{0.7820} \(\pm\) 0.0043 & 0.6446 \(\pm\) 0.0074 & 0.4973 \(\pm\) 0.0367 & 0.7284 \(\pm\) 0.0067 & \textbf{0.7960} \(\pm\) 0.0031\\
        \bottomrule
        \end{tabular}
    \end{adjustbox}
    \label{tab:result}
\end{table*}



Figure \ref{fig:model} provides a comprehensive overview of our model design.
The model begins with an embedding layer that encodes patient trajectories and timestamps into dense representations.
These embeddings are processed through stacked Transformer layers to capture temporal and relational dependencies, with the final output passed to a prediction layer for nowcasting laboratory test results.

\subsection{Medical Event Encoder}
Following standard practices for using Transformers with sequential data, we impose a restriction \(N\) on the maximum length of the input trajectory. Given an input patient trajectory \(\mathcal{T}_{p,v}=\{(e_i, t_i)\}^{N}\), we first calculate the hidden representation \(\mathbf{h}_i\) for each medical event \((e_i,t_i)\).

\textbf{Medical Code Embedding.}
Each code \(e_i\) is mapped to a dense vector representation $\mathbf{s}_i$ using a learnable embedding matrix \(\mathbf{E}_f\in\mathbb{R}^{\left|E\right|\times d}\), where \(\left|E\right|\) is the total number of medical codes and \(d\) is the embedding dimension. In addition, in order to capture the order information of events, we inject a learnable positional encoding $\mathbf{p}_i\in\mathbb{R}^{1\times d}$, which provides location information and helps the model understand the relative temporal order of events, to the code embedding. Finally, we obtain the medical code embedding of \(e_i\) as:
\begin{sequation}
    \mathbf{c}_i = \mathbf{s}_i + \mathbf{p}_i.
\end{sequation}%

\textbf{Timestamp Embedding.}
First, we model the decay over time based on the timestamp \(t_i\) with a square operation, as more recent events are more influential for future predictions. The decay is represented by:
\begin{sequation}
    \mathbf{f}^{\textit{decay}}_i = \mathbf{W}_d\left(1 - \tanh\left(\left(\mathbf{W}_tt_i-\mathbf{b}_t\right)^2\right)\right)-\mathbf{b}_d,
\label{decay}
\end{sequation}
where \(\mathbf{W}_d \in \mathbb{R}^{d\times h}\) and \(\mathbf{W}_t \in \mathbb{R}^{h\times 1}\) are learnable weighted matrix, \(\mathbf{b}_d \in \mathbb{R}^{d}\) and \(\mathbf{b}_t \in \mathbb{R}^{h}\) are bias vectors. Next, to capture the periodic patterns in medical treatments, such as regular laboratory tests or medications with daily cycles, we apply sine and cosine transformations to \(t_i\), leveraging their cyclic nature. This periodicity is represented by:
\begin{sequation}
    \mathbf{f}^{\textit{periodic}}_i = \mathbf{W}_p \cdot \begin{bmatrix} \sin\left( 2\pi\frac{t_i}{\omega} \right) \\ \cos\left( 2\pi\frac{t_i}{\omega} \right) \end{bmatrix} + \mathbf{b}_p,
\label{period}
\end{sequation}
where \(\mathbf{W}_p \in \mathbb{R}^{d \times 2}\) is a learnable weight matrix, \(\mathbf{b}_p \in \mathbb{R}^d\) is the corresponding bias term, and \(\omega\) represents the periodic time (e.g., for daily periodicity, \(\omega = 24\)). The decay and periodic components are combined via element-wise addition to form the final timestamp embedding:
\begin{sequation}
    \mathbf{f}_i = \mathbf{f}^{\textit{decay}}_i + \mathbf{f}^{\textit{periodic}}_i.
\label{fi}
\end{sequation}
By adding the medical code embedding and the timestamp embedding, we obtain the hidden representation of the input trajectory: 
\begin{sequation}
    \mathbf{h}_i = \mathbf{c}_i+\mathbf{f}_i.
\end{sequation}

\subsection{Time-aware Transformer}
Transformer layers iteratively calculate the hidden representation \(\mathbf{h}_i^l\) for the \(i\)-th event at layer \(l\), effectively encoding the patient's trajectory to capture temporal and contextual dependencies.
We first employ a multi-head self-attention mechanism to capture the complex relationships between medical events. Let \(h\) denote the number of heads and \(\mathbf{H}^l \in \mathbb{R}^{n \times d}\) denote the matrix of input embeddings obtained by stacking the input sequence of \(\mathbf{h}_i^l\). The multi-head attention is computed as:
\begin{align}
 &\operatorname{MultiHead}(Q, K, V)=\operatorname{Concat}\left(\operatorname{head}_1, \ldots, \text {head }_{h}\right) \mathbf{W^O}, \\
 &\text {head}_{i}=\text{softmax}\left(\frac{\mathbf{Q}^l{\mathbf{K}^l}^T\odot \mathbf{Z}^l}{\sqrt{d/h}}\right)\mathbf{V}^l, \label{attn}\\
& \mathbf{Q}^l=\mathbf{H}^l \mathbf{W}_i^Q, \mathbf{K}^l=\mathbf{H}^l \mathbf{W}_i^K,\mathbf{V}^l =  \mathbf{H}^l \mathbf{W}_i^V.
\label{attention}
\end{align}
Here, \(\mathbf{W}_i^Q\in\mathbb{R}^{d\times d / h}\),\(\mathbf{W}_i^K\in\mathbb{R}^{d\times d / h}\), and \(\mathbf{W}_i^V\in\mathbb{R}^{d\times d / h} \) are projection matrices for each head. \(\mathbf{W}^O\in\mathbb{R}^{d\times d} \) is the aggregation matrix. \(\mathbf{Z}^l\) is the denoising matrix. Sequential medical data has temporal irregularities and varying event importance. Additionally, noise like inconsistent time gaps or redundant records can affect performance. To address this, we apply a smoothed mask to denoise time-aware attention. For each layer \(l\), we define a mask matrix \(\mathbf{Z}^l\in\mathbb{R}^{N\times N}\), where \(N\) is the length of input trajectory. \(\mathbf{Z}^l\) contains logits that represent the relative importance between each query and key. Unlike binary masks \cite{10.1145/3523227.3546788, bastings-etal-2019-interpretable}, we use a smoothed mask matrix for more gradual modifications and easier optimization via gradient descent to improve denoising. \(\mathbf{Z}^l\) is randomly initiated with high positive values, and a loss function penalizes excessive logits in \(\mathbf{Z}^l\) to ensure effective denoising. Formally, the denoising loss is defined by:
\begin{sequation}
\mathcal{L}_{\text{denoise}}=\sum^L_{l=1}\lVert\mathbf{Z}^l\rVert_2.
\label{loss2}
\end{sequation}%

Following the attention mechanism, a position-wise feed-forward network (FFN) introduces non-linearity and transforms input features to capture dependencies beyond attention. Each sublayer (Figure \ref{fig:model}) includes residual connections and layer normalization to enhance performance. We stack such Transformer layers to capture complex patterns and hierarchical representations. 

\subsection{Masked Representation Prediction}

We adopt a masked representation prediction mechanism by appending a mask token \((e_{\text{mask}}, t_\text{mask})\), at the end of the sequence as a placeholder for the patient’s future state (Figure \ref{fig:model}). The model predicts the next clinical representation using the hidden representation of the mask event \(\mathbf{h}^L_\text{mask}\), processed through a linear layer with \textit{sigmoid} activation. The predicted probabilities $\hat{\mathbf{y}}$ are compared with the ground truth \(\mathbf{y}\) using cross-entropy loss:
\begin{sequation}
    \mathcal{L}_{\text{CE}}=-\frac{1}{\left|\mathcal{T}\right|}\sum^{\left|\mathcal{T}\right|}_{i=1}\left(\mathbf{y}_i\log(\hat{\mathbf{y}_i})+(1-\mathbf{y}_i)\log(1-\hat{\mathbf{y}_i})\right).
\end{sequation}
Combined with the denoising module’s loss, the final objective is:
\begin{sequation}
    \mathcal{L}_{\text{final}} = \mathcal{L}_{\text{CE}}+\lambda\mathcal{L}_{\text{denoise}}.
\end{sequation}

\section{Experiment}
  

\subsection{Experiment Setup}

{\textbf{Datasets.}} 
We evaluate our model on two real-world EHR datasets MIMIC-III \cite{Johnson2016MIMICIIIAF} and MIMIC-IV \cite{Johnson2023MIMICIVAF}.
They are randomly split into training, validation, and test sets, with a ratio of 0.75: 0.10: 0.15.

{\textbf{Baselines.}} We compare our model with the listed methods \textbf{(1)} Vanilla Transformer \cite{10.5555/3295222.3295349}, \textbf{(2)} RETAIN \cite{choi_retain_2016}, \textbf{(3)} Dipole \cite{ma2017dipole}, \textbf{(4)} HiTANET \cite{luo2020hitanet}, \textbf{(5)} ConCare \cite{ma2020concare}, \textbf{(6)} GRASP \cite{zhang_grasp_2021}, and \textbf{(7)} PPN \cite{yu_predict_2024}.

{\textbf{Evaluation Protocols and Metrics.}}
We evaluate our model using two protocols with distinct metrics. For multi-label classification, we use \textit{PR-AUC} and \textit{F1-score} to measure the model’s performance on imbalanced datasets. For ranking-based evaluation, we apply \textit{Precision@k} and \textit{NDCG@k} to assess its ability to rank relevant items at top positions. In \textit{NDCG@k}, ground truth events with identical timestamps are assigned the same rank.

\subsection{Experimental Results}
\begin{table}[tb]
\centering
\caption{Average count of model predictions. For simplicity, we selected a subset of baselines with relatively better performance. The threshold is set to 0.5.}
\begin{adjustbox}{max width=0.35\textwidth}
\begin{tabular}{@{}l|ccccc@{}}
\toprule
Model      & \textbf{MIMIC-III} & \textbf{MIMIC-IV} \\
\midrule
HiTANET           &   11670.86 \(\pm\)  483.92              &    2118.32 \(\pm\) 1223.93          &      \\
RETAIN           & 3098.37 \(\pm\) 179.53             & 2718.27 \(\pm\) 601.42            \\
PPN           & 3.95 \(\pm\) 0.68             & 14.49 \(\pm\) 0.52             \\
\model           & 25.44 \(\pm\) 1.29             & 20.49 \(\pm\) 1.06             \\
\midrule
Ground Truth   &   12.17 \(\pm\) 8.71      &        18.21 \(\pm\) 9.97\\
\bottomrule
\end{tabular}
\end{adjustbox}
\label{tab:avg_num}
\end{table}

\begin{table}[tb]
\centering
\caption{Ablation study results on MIMIC-III and MIMIC-IV.}
\begin{adjustbox}{max width=0.47\textwidth}

\begin{tabular}{@{}l|cccc@{}}
\toprule
\multirow{2}{*}{Model} & \multicolumn{2}{c}{MIMIC-III} & \multicolumn{2}{c}{MIMIC-IV} \\
\cmidrule(lr){2-3} \cmidrule(lr){4-5}
 & PR-AUC &Precision@5  & PR-AUC &Precision@5  \\
\midrule
w/o D      & 0.54(\(\downarrow 17.02\%\))   & 0.55(\(\downarrow 08.67\%\))    & 0.76(\(\downarrow 01.81\%\))&  0.75(\(\downarrow 01.47\%\))\\
w/o P      & 0.63(\(\downarrow 03.23\%\))  & 0.59(\(\downarrow 02.37\%\))   & 0.77(\(\downarrow 00.84\%\))&  0.75(\(\downarrow 00.38\%\))\\
w/o DP     & 0.48(\(\downarrow 25.98\%\)) & 0.53(\(\downarrow 12.61\%\))  & 0.73(\(\downarrow 06.26\%\)) &  0.74(\(\downarrow 02.58\%\))\\
w/o DPM    & 0.47(\(\downarrow 26.81\%\)) & 0.53 (\(\downarrow 13.31\%\))   & 0.72(\(\downarrow 06.69\%\))&  0.73(\(\downarrow 02.96\%\))\\
\midrule
\textbf{\model}        &  \textbf{0.65}  &  \textbf{0.61}   & \textbf{0.78} &  \textbf{0.76} \\
\bottomrule
\end{tabular}

\end{adjustbox}
\label{tab:ablation_study}
\end{table}

Table \ref{tab:result} presents the experimental results, comparing \model~with baseline models across two datasets and evaluating performance on various metrics to highlight strengths and limitations. Our analysis reveals key patterns and insights into how different models handle the newly defined laboratory measurement prediction task. Our analyses are threefold:

\textbf{(a)} The results on MIMIC-IV are generally better than on MIMIC-III, due to its richer, more comprehensive data, including medications and procedures. Unlike MIMIC-III, which focuses mainly on laboratory measurements, MIMIC-IV offers more detailed information including flags for test results, helping the model better understand the context and improve prediction accuracy.

\textbf{(b)} Transformer-based models outperform RNN-based models in handling long sequences with new input structures. HiTANET, leveraging the Transformer architecture, achieves the highest Precision, NDCG, and AUC-PR on MIMIC-III by incorporating temporal information. Baseline models originally designed for \textit{inter-visit} predictions struggle with handling event-level data. RETAIN’s linear embedding layer enables it to effectively model event sequences, outperforming those models.

\textbf{(c)} \model~ excelled across all evaluation metrics, outperforming other models on both MIMIC-III and MIMIC-IV. With Transformer architecture and a strong embedding strategy, \model~ adapts effectively to intra-visit prediction tasks and achieves outstanding predictive accuracy. For the multi-label classification task, \model~ also outperformed baseline models in predicting the correct number of laboratory measurements. As shown in Table \ref{tab:avg_num}, while baselines often over-or underestimated the number of measurements, \model~ aligned more precisely with the actual number of events.


\subsection{Abalation Study}
To evaluate key components of \model, we conducted an ablation study focusing on the decay feature (D), periodicity feature (P) of the timestamp embedding, and the denoising module (M). Experiment results are shown in Table \ref{tab:ablation_study}.



{\textbf{Without D/P.}}
The results show a significant performance decline when temporal features are excluded. Removing the decay feature led to a 17.02\% drop in PR-AUC and an 8.67\% drop in Precision@5 on MIMIC-III. Excluding the periodicity feature (w/o P) caused a 3.23\% decrease in PR-AUC on MIMIC-III and 0.84\% on MIMIC-IV, highlighting both features' importance.


{\textbf{Without DP.}}
When both decay and periodicity features were removed, PR-AUC dropped by 25.98\% and Precision@5 by 12.61\% on MIMIC-III, confirming their complementary roles in capturing temporal dynamics. Smaller but notable declines on MIMIC-IV further underscore their importance across datasets.


{\textbf{Without DPM.}}
The masking mechanism showed a notable impact. \model~ w/o DPM exhibited the worst performance across all metrics on both datasets. This highlights the essential role of the denoising module in filtering out irrelevant or noisy information, thereby enhancing the model’s robustness and predictive accuracy.

\subsection{Case Study}
\begin{figure}[t]
    \centering
    \includegraphics[width=0.87\linewidth]{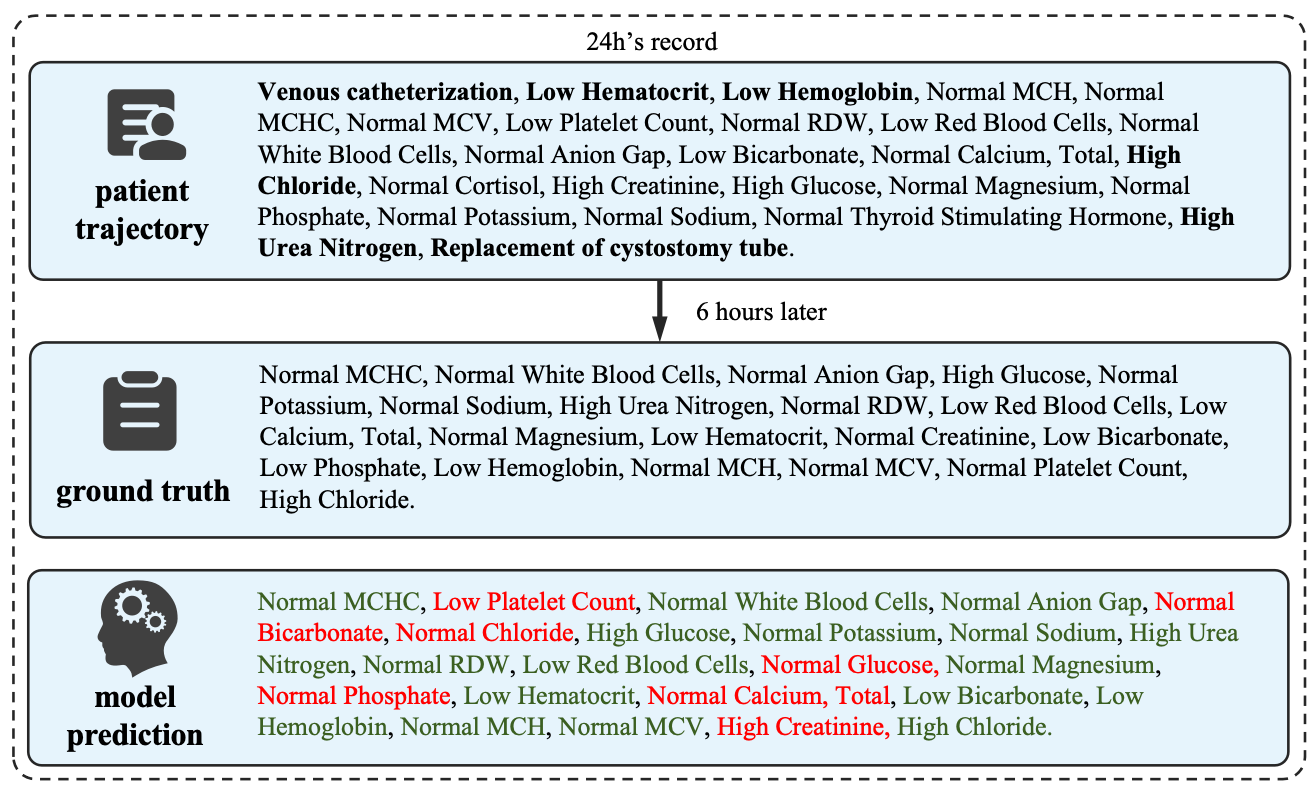}
    \caption{A sample case of a patient with kidney dysfunction. For simplicity, the ordered patient trajectory only contains several important medical events. The wrong predictions are colored with red and the correct ones are colored with green.}
    \label{fig:case}
\end{figure}

In this case study, we analyze a hospitalized patient’s trajectory and demonstrate how \model~ predicts critical laboratory measurements, aiding clinical decision-making. Figure \ref{fig:case} depicts a sample trajectory and \model’s predictions.

\textbf{Patient Background.}
The hospitalized patient underwent procedures like \textit{venous catheterization} and \textit{cystostomy tube replacement}, indicating fluid management and urinary tract care. Key laboratory tests showed elevated \textit{urea nitrogen} and \textit{chloride}, suggesting renal impairment and metabolic acidosis, along with low \textit{hematocrit}, \textit{hemoglobin}, and \textit{platelets}, indicating anemia.

\textbf{Model Prediction.}
\model~ accurately predicts a range of laboratory values from the patient’s prior trajectory, aligning well with the ground truth. It correctly identified a normal MCHC, low RBC, and high urea nitrogen,
but struggles with values like glucose and creatinine, likely due to limited 24-hour historical data.
Despite discrepancies, 
\model~helps clinicians prioritize management and guide treatment decisions.

\section{Conclusion and Future Work}
In conclusion, this work defines the novel task of laboratory measurement prediction within a hospital visit.
We propose \model, which successfully captures intricate temporal dependencies, enabling fine-grained predictions that reflect the dynamics of patient data. Our findings emphasize the value of \textit{intra-visit} information and demonstrate how processing long sequences enhances clinical decision-making accuracy and timeliness. Future work will focus on two directions: 
(1) Leveraging the external knowledge~\cite{li2024retrievalreasoningrerankingcontextenriched,li2024contextawareinductiveknowledgegraph} and/or Large Language Models with particular safety and privacy considerations~\cite{li2024ultrawiki,liu2024an,liu2024a} for knowledge adaptation and EHR data parsing~\cite{qiu2024ease}.
(2) Enhancing the model by integrating advanced temporal learning and trajectory analysis techniques to further improve its accuracy, robustness, and continual learning capability~\cite{zhang2024influential,yu2024recent,rossi2020temporal,han2020traffic,han2022deeptea,han2024fdm}. 

\begin{acks}
    This work was supported in part by NSFC (Grants 62302421, 62102341), the Guangdong Talent Program (Grant 2021QN02X826), the Basic and Applied Basic Research Fund in Guangdong Province (Grant 2023A1515011280), the Shenzhen Science and Technology Program (Grants JCYJ20220530143602006, ZDSYS 20211021111415025), and the Shenzhen Research Institute of Big Data (Grants SIF20240004, SIF20240002). Additional support was provided by the CCF-Ant Research Fund, the Guangdong Provincial Key Laboratory of Big Data Computing at the Chinese University of Hong Kong, Shenzhen, and the Natural Sciences and Engineering Research Council of Canada (Grant RGPIN-2020-05408).
\end{acks}

\bibliographystyle{ACM-Reference-Format}
\bibliography{EHR}
\end{document}